\documentclass{article}

\usepackage{PRIMEarxiv}
\usepackage{amsmath}
\usepackage[utf8]{inputenc}
\usepackage[T1]{fontenc}
\usepackage{hyperref}
\usepackage{url}
\usepackage{booktabs}
\usepackage{amsfonts}
\usepackage{nicefrac}
\usepackage{microtype}
\usepackage{lipsum}
\usepackage{fancyhdr}
\usepackage{graphicx}
\graphicspath{{media/}}
\usepackage{tikz}

\title{Multi-Agent Collaboration: Harnessing the Power of Intelligent LLM Agents}

\author{
  \href{https://orcid.org/0009-0002-5792-6464}{\includegraphics[scale=0.06]{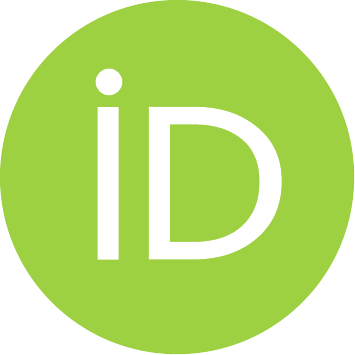}\hspace{1mm}Yashar Talebirad} \\
  University of Alberta \\
  Edmonton, Alberta, Canada\\
  \texttt{talebira@ualberta.ca} \\
   \And
  \href{https://orcid.org/0000-0003-4112-2138}{\includegraphics[scale=0.06]{orcid.pdf}\hspace{1mm}Amirhossein Nadiri} \\
  York University \\
  Toronto, Ontaria, Canada\\
  \texttt{anadiri@yorku.ca} \\
}

\begin{document}
\maketitle

\begin{abstract}
In this paper, we present a novel framework for enhancing the capabilities of large language models (LLMs) by leveraging the power of multi-agent systems. Our framework introduces a collaborative environment where multiple intelligent agent components, each with distinctive attributes and roles, work together to handle complex tasks more efficiently and effectively. We demonstrate the practicality and versatility of our framework through case studies in artificial general intelligence (AGI), specifically focusing on the Auto-GPT and BabyAGI models. We also examine the "Gorilla" model, which integrates external APIs into the LLM. Our framework addresses limitations and challenges such as looping issues, security risks, scalability, system evaluation, and ethical considerations. By modeling various domains such as courtroom simulations and software development scenarios, we showcase the potential applications and benefits of our proposed multi-agent system. Our framework provides an avenue for advancing the capabilities and performance of LLMs through collaboration and knowledge exchange among intelligent agents.
\end{abstract}

\section{Introduction}

The field of artificial intelligence is rapidly advancing, bringing with it the complexity and challenges of deploying AI models to manage an array of tasks. In response to these challenges, researchers are delving into multi-agent systems where multiple AI entities collaborate towards a common goal \cite{dohan2022language}. 
One such multi-agent system can be seen in the work of \cite{park2023generative}, who introduced generative agents that imitate plausible human behavior within an interactive sandbox environment. 
Another instance of this exploration is Camel \cite{li2023camel}, in which a system is introduced which leverages a Large Language Model (LLM) to generate diverse and detailed instructions for a wide range of tasks. It incorporates role-playing scenarios where two agents interact, thereby demonstrating the potential of such systems in handling complex real-world scenarios.
In a similar vein, our paper proposes the use of multiple LLMs, each bearing diverse characteristics, to enhance performance across a range of tasks. 

We focus particularly on the recent iterations of the Generative Pretrained Transformer (GPT) models, GPT-4 and GPT-3.5-turbo. From content creation and question-answering systems to language translation, GPT models have manifested immense potential across a plethora of applications. Early experiments with GPT-4 \cite{bubeck2023sparks} reinforces this fact, showing GPT-4's ability to solve complex tasks across diverse fields, approaching human-level performance. As a result, the adeptness of these models at managing complex language tasks makes them ideal candidates for our purposes. Moving forward, we will use the term ``Intelligent Generative Agents'' (IGAs) to refer to a series of agents, each equipped with unique attributes and roles. GPT models, despite their commendable text generation capabilities, operate as isolated entities in their conventional form. They lack the capability to collaborate with other agents or draw from external knowledge repositories. This inherent shortcoming restricts their utility in complex scenarios that necessitate collaborative efforts and information sharing among multiple intelligent systems.

Our proposed use of multiple IGAs stems from the notion that diversity in a system enhances performance. The idea is based on the principle of division of labor, where each agent specializes in a specific function, thereby improving the efficiency and effectiveness of the system as a whole. This mirrors the concept of teamwork in human systems, where different individuals contribute their expertise to complete a task. A diverse set of agents, each configured with unique strengths, can collectively handle a wider range of potential inputs, outputs, and processing strategies. Furthermore, delegating different roles to each agent introduces more flexibility and efficiency in the context task management. This brings about the concept of ``multi-agent systems'', in which numerous IGAs interact and collaborate to achieve a shared goal. These agents are capable of creating subtasks, seeking information, and soliciting assistance from each other, and can also engage in competitive evaluation for better outcomes. The emphasis on collaboration and knowledge exchange in multi-agent systems can bolster the problem-solving proficiency of GPT models, thereby paving the way for more sophisticated intelligent systems. In fact, our proposed multi-agent system also aims to make strides toward achieving a higher level of artificial general intelligence (AGI). The development and deployment of advanced generative AI models like GPT-4, represent significant steps towards AGI \cite{zhang2023small}. By fostering collaboration and knowledge exchange among multiple IGAs, our system seeks to further this progress. It aims to emulate the diverse and adaptable problem-solving capabilities that are characteristic of an AGI, thereby pushing the boundaries of what AI can achieve.

Our proposed abstraction allows users to engage with a ``black box'' by providing an initial prompt and receiving the final output without grappling with the underlying complexities of agent collaborations and interactions. Consider, for instance, taking inspiration from the success of Generative Adversarial Networks (GANs), where two networks (a generator and a discriminator) work collaboratively to produce better results, a simple yet effective multi-agent system can be made utilizing two IGAs: One with memory and one with Internet access. By combining their strengths, these agents could cooperate to reduce the occurrence of 'hallucinations' or false information generation, thereby increasing the reliability of the output \cite{bang2023multitask}. This could be particularly beneficial in tasks where accuracy and fact-checking are critical. 

The objectives of this paper are to explore and demonstrate the potential of having multiple agents within a black box environment for enhanced collaboration and problem-solving. The specific objectives are outlined as follows:

\begin{itemize}
    \item Introducing a General Framework: The primary objective is to pave the way for the creation of more powerful AGI models. By providing a general framework for multi-agent systems using LLMs, we aim to push the boundaries of what AI can achieve, thereby contributing to the advancement toward Artificial General Intelligence.
    \item Adaptive and Dynamic System: We introduce a system that is adaptive and can change itself to suit the tasks at hand. Having a static structure will limit a system to a set of predefined tasks. The possibility of addition and removal of agents in the system will make it flexible and capable of solving more complex problems.
    \item Multi-Agent Collaboration: In this paper, we explore the use of multiple LLMs in a collaborative manner. This collaboration aims to enhance performance across diverse tasks, utilizing the strengths of each agent and encouraging a synergistic relationship amongst them.
\end{itemize}

By effectively addressing these objectives, this paper aims to significantly advance the understanding and utilization of multi-agent collaboration in the realm of IGAs. The exploration and demonstration of such a model of collaboration serve as a stepping stone for future research in this domain.

The remainder of this paper is organized as follows: We begin by laying the foundation of our discussion in Section \ref{environment}, where we introduce the essential components that make up the proposed framework. Following this, Section \ref{Overview} provides a comprehensive description of the proposed multi-agent framework and its functionalities. In Section \ref{application}, we explore various potential use cases and applications of this framework, demonstrating its versatility and adaptability. Section \ref{challenges} then discusses the potential challenges associated with our approach, shedding light on its limitations. We conclude our discussion in Section \ref{conclusion}, where we summarize the main points of our discussion, draw final conclusions, and suggest directions for future research in this domain.

\section{Building Blocks of the Multi-Agent System} 
\label{environment}

The environment in which the multi-agent system operates can be considered a black box. This is due to the inherent nature of IGAs, which are trained on vast amounts of data and generate outputs based on complex internal computations that are not directly observable or interpretable. This black box environment is a digital workspace where multiple instances of IGAs interact and collaborate. This workspace provides a shared platform for the agents to communicate, exchange information, and perform tasks. Additionally, plugins can be used to provide additional functionality or capabilities to agents. They can be seen as a specialized tool or service that agents can utilize to perform specific tasks or access certain (internal or external) resources. Any of the agents or plugins can also be responsible with receiving the initial prompt from the user or responding to the main user.

We denote the black box environment as a graph $G(V, E)$, where $V$ is the set of vertices representing the IGAs and the plugins, and $E$ is the set of edges representing the connection channels between the agents and the plugins, and between the agents themselves.

\subsection{Agent Representation}

Each agent $i \in V$ is represented as a tuple $A_i = (L_i, R_i, S_i, C_i, H_i)$, where:

\begin{itemize}
\item $L_i$ refers to the language model instance utilized by the agent. This encompasses the model's type (such as GPT-4 or GPT-3.5-turbo) and its specific configuration parameters, including the 'temperature' setting which influences the degree of randomness in the agent's output. The choice of the language model can be dictated by the task requirements. For instance, while GPT-4, due to its exceptional reasoning capabilities, could be assigned tasks demanding deep insights and complex problem-solving, GPT-3.5-turbo could be employed for tasks requiring quicker execution owing to its faster performance.
\item $R_i$ is the role of the agent. The role defines the responsibilities of the agent and provides the agent with a sense of purpose and direction, guiding its actions and interactions. More specifically, these responsibilities are the tasks or functions the agent is expected to carry out within the system. For instance, an agent's responsibilities could include processing and responding to user queries, coordinating interactions between other agents, managing resources, or overseeing a particular aspect of the system's operations.
\item $S_i$ is the state of the agent, encompassing its current knowledge base and its "thoughts". The agent's state evolves over time based on the tasks it performs and the interactions it has with other agents or the environment. 
\begin{itemize}
\item The ``knowledge'' component of the state can be seen as a representation of the agent's current understanding or awareness of its environment, tasks, and interactions. It's updated whenever the agent learns new information or gains a new understanding.
\item ``Thoughts'' in this context can be interpreted as the agent's current focus or intent \cite{wei2023chainofthought}. They represent what the agent is currently contemplating or planning, based on its knowledge and the task at hand. Thoughts may inform the agent's next action and may be updated after each operation. They may also encapsulate the agent's internal dialogue or reasoning process as it works towards its goal.
\end{itemize}
\item $C_i$ is the boolean property indicating whether an agent has the ability to create new agents. By setting this property to true, an agent can dynamically generate new agents within the system. 
\item $H_i$ is the set of agents that this agent has the authority to halt. By specifying the agents that can be halted, an agent can exercise control over the execution of other agents within the system. 
\end{itemize}

\subsection{Plugin Representation}

Each plugin $j \in V$ is represented as a tuple $P_j = (F_j, C_j, U_j)$, where:

\begin{itemize}
\item $F_j$ is the set of the functionalities of the plugin, which are the actions that an agent can perform. This can include accessing and manipulating digital resources within this environment, such as files and databases, and interacting with external systems through APIs and other interfaces.
\item $C_j$ are the configurations associated with the plugin. Parameters are variables that are set when the plugin is initialized. Examples include API keys for accessing specific services, or threshold values to determine specific behavior. Parameters help in customizing the functionality of the plugin according to the task or application at hand.
\item $U_j$ is the set of usage constraints or conditions that govern the usage of the plugin. These constraints define the boundaries and capabilities of the plugin and can include limitations in terms of computational resources, input data types, and output capabilities. 

\end{itemize}

\subsection{Connection and Message Representation}

Each edge $e_{ij} \in E$ connects an agent $A_i$ to either a plugin $P_j$ or another agent $A_j$ using a communication channel. Each agent can send messages through the channels that it is connected to, and each message $m \in M_{ij}$, sent from agent $A_i$ to $A_j$, is represented as a tuple $m = (S_m, A_m, D_m)$, where:

\begin{itemize}
\item $S_m$ is the content of the message.
\item $A_m$ is the action associated with the message, which can be a task assignment, a report, a request, or any other action.
\item $D_m$ is the metadata associated with the message, which can include information such as the timestamp, sender, receiver, and potentially additional context-specific data.
\end{itemize}

Another approach to data transmission between agents can involve the use of plugins. For example, plugins designed for data storage can serve as shared databases, enabling different agents to access and retrieve information stored by other agents. Further extending this concept, a plugin could act as a communication board, enabling multi-directional communication between multiple agents. This essentially forms a many-to-many communication platform within the system.

\section{Detailing Proposed Framework}
\label{Overview}
In any multi-agent system, the nature of interaction and collaboration between the agents play a significant role in determining the overall system performance. This section explores the ways in which these interactions can be managed and optimized, particularly in the context of a system composed of multiple IGAs.

\subsection{System Design}

The design of a multi-agent system involves determining the number of agents, the required plugins, establishing connections between agents and plugins, creating connections between agents to enable communication, and assigning roles and properties of agents. This design aims to optimize the configuration and align it with the desired end goal of the system, enabling efficient collaboration and interaction among the agents.
 
While designing the system, the following steps are taken:
\begin{itemize}
    \item Agent Roles: Roles for the agents are identified and defined within the environment, based on the specific requirements of the task at hand. Each agent is assigned a role, which specifies their responsibilities and duties in the system.
    \item Agent-Plugin Connections: Connections between agents and plugins are established to provide agents with additional functionality. By connecting agents to plugins, agents gain access to tools, resources, or external services that enhance their capabilities. These connections allow agents to leverage the functionalities of the plugins.
    \item Agent-Agent Connections: Connections between agents are created to enable communication and collaboration. These connections allow agents to exchange messages, share information, and cooperate toward achieving the common goal.
    \item System Operations: Agents can be granted specific permissions to create new agents or halt a specific set of agents. Also, any of the plugins or agents can be responsible for receiving the initial prompt from the user or responding to them. 
\end{itemize}

By carefully designing the system with well-defined agents, plugins and the connections between them, the framework enables efficient multi-agent interaction and collaboration. This design allows agents to effectively communicate, coordinate, and work together towards achieving the common goal within the black box environment.

\subsection{Dynamic Addition of Agents}

In certain scenarios, an agent with the ability to create new agents may dynamically add additional agents to the system. This capability enables agents to distribute their workload and assign specific responsibilities to enhance collaboration and workload management. This need may arise as a byproduct of a sudden increase in the workload of the system. When a new agent is created, the creator assigns the new agent a role, grants it the necessary properties, and establishes connections with other agents and plugins. These properties and connections are subsets of those available to the creator agent. Also, a connection to the creator is established.

Once the new agent is created and initialized, it operates independently within its defined role. The creator agent sets a clear goal for the new agent, providing initial guidance to ensure a smooth transition of responsibilities. By allowing agents to dynamically create new agents and delegate tasks, the system can effectively manage workloads, enhance parallel processing capabilities, and improve overall system performance. This dynamic approach fosters a collaborative environment where agents can dynamically organize and distribute tasks, ultimately contributing to the achievement of the common goal.

The fact that a designer designed the system and defined the capabilities, connections, and permissions of the agents does not contradict the dynamic addition of agents and their ability to distribute workload and delegate responsibilities. Although the designer has designed the initial framework, the dynamic addition of agents allows for flexibility and adaptation within the designed system. It empowers the agents themselves to make decisions and create new agents based on their own assessments of workload and the need for assistance. The designer's role is to provide the initial structure and guidelines, but the system allows for agent autonomy and self-organization.

Hence, system design and the dynamic addition of agents function harmoniously. The initial framework laid out by the designer serves as a robust foundation, while the agents' ability to dynamically adapt and distribute workload ensures flexibility and resilience under changing conditions and demands.

\subsection{Feedback and Self-Feedback Mechanisms}

Feedback mechanisms play a pivotal role in multi-agent systems, enabling agents to learn from their experiences and adapt their strategies for improved performance. These mechanisms can be categorized into inter-agent feedback and self-feedback \cite{fu2023improving, chen2023teaching, madaan2023selfrefine}. Inter-agent feedback involves agents providing constructive criticism to each other based on their interactions and collaborations. Such feedback can help agents identify areas of improvement and adapt their strategies accordingly, enabling continuous learning and improvement within the system \cite{fu2023improving}. Some multi-agent systems employ inter-agent feedback by involving agents into role-playing. This approach involves designing specific prompts (denoted as Inception Prompting in \cite{li2023camel}) to guide chat agents toward task completion while maintaining consistency with the main goal. This approach can be integrated in the proposed model, giving different roles to multiple agents and connecting them together.

Self-feedback, on the other hand, involves agents assessing their own performance and identifying areas of improvement. This can be achieved through a self-assessment mechanism where agents evaluate their performance based on predefined criteria or goals. This self-assessment can help agents identify their strengths and weaknesses, adapt their strategies, and improve their problem-solving capabilities \cite{chen2023teaching}. In the proposed model, self-feedback can be simulated by a pair of connected agents: one with the role of giving feedback and the other tasked with refining the response based on the feedback received. Note that this simulation removes the need for a human to ask for possible refinement of the response.

\subsection{Oracle Agent}

An oracle agent is a unique type of agent in the system that operates in a stateless and memory-less manner. Unlike other agents that may maintain a state or memory to guide their actions, an oracle agent performs actions based solely on the current input it receives, without any regard for past inputs or outputs. This characteristic makes oracle agents particularly useful in scenarios where the task at hand is independent of previous interactions or states.

Every interaction with an oracle agent is treated as an isolated event, independent of any previous interactions. This makes oracle agents highly predictable, as their actions are solely determined by the current input and not influenced by any past events. oracle agents are mainly designed to be utilized by other agents. For instance, an oracle agent can give feedback on the response of the other agents, and let them refine their responses before proceeding.

\subsection{Halting Mechanism and Supervision}
\label{halting}

The proposed framework incorporates an essential mechanism whereby an agent can halt other agents under certain conditions. This capability is crucial for effective management and coordination of tasks within a multi-agent system. Specifically, this ability can be granted to any agent in the system, including those that create new agents. The authority to halt becomes inherently necessary for these parent agents to maintain control and ensure the proper functioning of their created agents.

In practice, an agent halting another would involve signaling the targeted agent to cease its current activity. This signaling could be in the form of a command or a message transmitted via the communication interfaces defined within the system. Upon receiving this signal, the halted agent would immediately stop its current operation and wait for further instructions. Depending upon the system design, it could either enter an idle state or undertake a default operation in such instances. For creator agents and the agents they created, the halting mechanism works similarly. If a creator agent identifies undesirable activity in its created agent, it can initiate the halt command, causing them to stop their current operation immediately. This interaction emphasizes the supervisory role of the creator agent, ensuring that created agent functions correctly and does not deviate from its intended role.

In fact, this supervisory role can be enhanced by the introduction of a specialized ``Supervisor Agent''. This Supervisor Agent can monitor the progress and task list of the main agent, providing timely feedback when necessary. For example, if an agent is stuck in a loop or deviates from its assigned task, the Supervisor Agent can detect these issues by reviewing recent activities. Upon such detection, the Supervisor Agent can initiate the halt command, prompting the main agent to cease its current operation and change its course of action. This mechanism not only facilitates better task management but also reduces the need for constant human monitoring and intervention in the feedback process.

\subsection{Autonomous System Design}

One notable aspect of the proposed framework is the potential role of an intelligent LLM as the system designer. The unique capabilities of an IGA extend beyond being just an agent within the environment, as it possesses the ability to fulfill the role of designing the system itself. It can consider the system's objectives, constraints, and desired functionalities to define the roles and responsibilities assigned to each agent. Additionally, the IGA can employ its knowledge of communication protocols and collaborative frameworks to determine the optimal interactions and connections between agents, plugins, and other system components. Drawing upon its comprehensive understanding of the problem domain, combined with precise system formulation and specified objectives, the IGA can design an effective system that optimally addresses the task at hand.
Alternatively, after a human designs the initial system, an IGA can analyze the system structure, roles, interactions, and connections, and provide feedback and refinement to an already designed system. The IGA can also utilize its natural language generation capabilities to communicate the system design to the system owners. It can provide clear and concise descriptions of the agents' positions, roles, and interactions, allowing for a comprehensive understanding of the system's structure and functioning.

\section{Use Cases and Applications}
\label{application}
This section aims to demonstrate the practicality and versatility of the proposed multi-agent framework by examining its applicability to existing AI models. We focus specifically on two cutting-edge artificial general intelligence (AGI) models, Auto-GPT\footnote{\href{https://github.com/Significant-Gravitas/Auto-GPT}{https://github.com/Significant-Gravitas/Auto-GPT}} and BabyAGI\footnote{\href{https://github.com/yoheinakajima/babyagi}{https://github.com/yoheinakajima/babyagi}}, and examine how our framework could potentially enhance their design and operation. We explore the models' main components, their operation, and limitations, and how our framework could be applied to improve their performance. Additionally, we discuss potential modifications that our framework can add, thus offering a broader understanding of the potential applications and benefits of the proposed multi-agent system.

\subsection{Artificial General Intelligence: Auto-GPT}

Auto-GPT is an experimental open-source AI application that has been gaining significant attention due to its promising autonomous abilities. It is considered a step towards AGI, a type of AI that can perform human-level intellectual tasks. Auto-GPT has internet access, long-term and short-term memory management, GPT-4 for text generation, and file storage and summarization with GPT-3.5. It can perform tasks that ChatGPT \cite{chatgpt} can do, such as debugging code and writing an email, but it can also complete more advanced tasks with fewer prompts. Auto-GPT's design is based on the concept of thoughts, which are essentially the steps it takes to complete a task. 

\subsubsection{Model}
The framework on which Auto-GPT runs can be modeled using our proposed framework.
We can consider the Auto-GPT's main agent as a single agent in our model.
The agent's goal is to perform tasks autonomously by chaining thoughts together, while working towards the goals specified by the user. The state of the agent includes the current task it is working on, and the chain of thoughts it has generated so far.
This agent can also create other agents and halt any of them.
Plugins can be represented as external services or tools that the agent uses to perform its tasks. For example, browsing the internet, managing memory, interacting with files, executing code, generating images, and similar tasks can be identified as plugins in our framework. 
There will also be an oracle agent, which is responsible for tasks such as summarization and criticizing the responses of the main agent.
These plugins, along with the agents the main agent creates, can all be considered as nodes in the graph corresponding to the system, and the connections between the Auto-GPT agent and its plugins, along with the connections between the agent and the other agents it makes, can be represented as edges in the graph. 
Messages sent through these connections may include task assignments, requests for information, or commands to execute certain operations. 
\subsubsection{Limitations and Possible Improvements}
Despite its potential, Auto-GPT faces several challenges and limitations. One significant obstacle is that it might get stuck in a loop, rendering it unable to function properly. 
The looping issue is a result of the system's reliance on chaining thoughts together to perform tasks. While this approach allows the system to perform complex tasks, it also makes it prone to getting stuck in loops, especially when dealing with complex or ambiguous problems. However, features that are proposed in our framework can possibly address this shortcoming, and open further avenues for improvements. For instance, the agent's inability to realize when it has got stuck or notice that it has gone off task can potentially be mitigated by adding the ``Supervisor Agent'' that was introduced in Section \ref{halting}.

As another example, one can implement a concept of co-agents, where multiple autonomous instances of Auto-GPT could collaborate, share a workspace for files, and communicate in a board, essentially mimicking a team of humans working remotely, with each having a specific role. 

Additionally, the system's ability to interact with files and execute code opens up a wide range of possibilities for its use, but it also introduces potential security risks. These risks are possibly alleviated by having a human provide feedback and authorizing each step, but this step is completely ignored when using the app's ``continuous'' mode. This means that the system should be designed with robust security measures in place to prevent unauthorized access or misuse. This can be done using a state-less oracle agent, which can monitor each sensitive task and decide if it is indeed malicious or not.

\subsection{Artificial General Intelligence: BabyAGI}

BabyAGI is an AI agent that can generate and attempt to execute tasks based on a given objective. BabyAGI operates based on three LLM chains: Task creation chain, Task prioritization chain, and Execution chain. 

\subsubsection{Model}
In our proposed framework, BabyAGI can be modeled as a system of interconnected agents, each with a specific role. The agents in BabyAGI include a task creation agent, a task prioritization agent, and an execution agent. 
In addition to these agents, BabyAGI uses a vector database to store and retrieve task results for context. 
This can be modeled as a plugin in our framework that interacts with a vector database, with operations for storing and retrieving data. Furthermore, there can be an additional agent in our framework that interacts with the user, refines the input, and places it into the task storage.

\subsubsection{Limitations and Possible Improvements}

Our framework can potentially improve upon the current implementation of BabyAGI by providing a more structured and modular approach to designing the system. By modeling each agent, plugin, and operation explicitly, our framework can make it easier to understand and modify the system. Furthermore, our framework's support for feedback loop can enable the agents in BabyAGI to learn from their own performance and improve over time.

\subsection{The ``Gorilla'' Model}

\subsubsection{Model}

The Gorilla system \cite{patil2023gorilla} is based on a fine-tuned LLaMA \cite{touvron2023llama} model with additional capabilities to retrieve documents and integrate this information during both training and inference. It is capable of extending beyond mere language modelling, embracing features that enable interaction with external APIs, handling document retrieval, and adaptation to version changes. In this system, API calls and their documentation are used to instruct the LLM about the specific tasks each API can handle. The model learns to map prompts to API calls by using a retrieval system to access the most up-to-date API documentation from the database. Gorilla also mitigates hallucination and adapts to changes in API documentation by using a retriever during training and inference.

In our framework, we need a single agent to model the Gorilla system. To handle APIs, our model can employ plugins, which can be seen as extensions or modules designed to handle specific tasks.
This results in enhanced flexibility and versatility in the system that allows it to handle a broader range of tasks.

\subsubsection{Limitations and Possible Improvements}

Although integrating information during both training and inference shows significant improvements over GPT-4 in writing API calls, our model offers a more generalized and robust framework that can be customized to different use cases. For example, our model can handle real-time updates to APIs and their documentation more efficiently by updating the relevant agent's knowledge base, rather than having to update the entire model. Additionally, it can handle overlapping functionality between different APIs more elegantly by deciding between different agents based on their functionality.

Our model can also potentially improve the process of mitigating hallucinations by using a dedicated agent for this task. This agent could verify the main agent's responses to find out when the agent is hallucinating and intervene to correct the output.

Our model can further improve the process of interacting with APIs by employing different agents for different APIs, each equipped with its plugin for the relevant API documentation. 
This would allow our model to handle more complex tasks and interactions, as it can leverage the combined capabilities of multiple agents at once.

\subsection{Case Study}

In this section, we will delve into two distinct case studies to illustrate the practical applications of our proposed multi-agent system. These case studies, namely a court simulation and a software development scenario, have been chosen due to their inherent complexity and the necessity for diverse roles and interactions within them. Both scenarios involve a multitude of tasks and responsibilities that need to be coordinated effectively for successful outcomes. By employing our multi-agent framework, we aim to demonstrate how such complex processes can be modeled in a common framework. Each agent in the system will be assigned a specific role, mirroring the real-world roles in these scenarios. They will be equipped with the necessary tools and capabilities to fulfill their responsibilities, thereby contributing to the overall objective. 

\subsubsection{Court Simulation}

Before the introduction of new LLMs, attempts to simulate environments like a courtroom required training with specific data \cite{hamilton2023blind}. However, with the recent advancements in the area of language models, the training process might not be necessary anymore. In this context, our framework can be utilized to model the various roles and interactions that take place within a courtroom. This includes the roles of the judge, jury, attorneys, witnesses, and even the court clerk. Each of these roles can be represented by an agent within the system, with specific responsibilities and capabilities assigned to them.

\begin{itemize}
\item \textbf{Judge Agent:} The Judge Agent is responsible for overseeing the proceedings, making rulings on legal issues, and ultimately delivering the verdict in non-jury trials. This agent would require a plugin that provides access to a comprehensive database of legal knowledge and precedents, enabling it to make informed decisions.

\item \textbf{Jury Agent:} The Jury Agent is responsible for determining the facts of the case and delivering a verdict in jury trials. This agent would require a plugin that allows it to understand and evaluate the evidence presented during the trial. 

\item \textbf{Attorney Agents:} The Attorney Agents represent the prosecution and the defence in the trial. They are responsible for presenting their respective cases, cross-examining witnesses, and making closing arguments. These agents would require plugins that provide access to legal knowledge, as well as plugins that enable them to understand and generate persuasive arguments.

\item \textbf{Witness Agents:} The Witness Agents provide testimony during the trial. They would require plugins that allow them to accurately recall and describe events.

\item \textbf{Court Clerk Agent:} The Court Clerk Agent is responsible for administrative tasks such as maintaining court records and administering oaths. This agent would require plugins that enable it to manage and retrieve documents, as well as plugins that allow it to perform its administrative duties.
\end{itemize}

The interactions between these agents would be governed by a set of predefined rules and protocols, simulating the procedures followed in a real courtroom. For instance, the Judge Agent could issue instructions to the other agents, the Attorney Agents could question the Witness Agents, and the Jury Agent could request clarification or additional information from any of the other agents.

In terms of operations, the simulation process would proceed in stages, similar to a real trial. The Attorney agents would present their opening statements, followed by the presentation of evidence and witness testimony. The Jury Agent would then deliberate and deliver a verdict, after which the Judge Agent would conclude the proceedings.

This simulation could be used for a variety of purposes, such as training for law students, testing new legal theories, or even automating certain aspects of the legal process. However, it's important to note that while our framework can simulate the process and interactions in a courtroom, it cannot fully replicate the complexities of human decision-making and judgement. Therefore, any outcomes produced by the simulation should be interpreted with caution.

\subsubsection{Software Development}

Our model can be effectively used in the context of software development, enabling the creation of a multi-agent system where each agent embodies a specific role integral to the software development process. By assigning distinct responsibilities to individual agents, the development process can be significantly optimized and streamlined. The key roles, as derived from the software development team structure, can be represented as follows:

\begin{itemize}
    \item \textbf{User Experience Designer}: This agent is responsible for understanding and designing the user experience. It can use a plugin that simulates user interactions to test different designs and gather data on user preferences. The agent can then use this data to refine the design.
    \item \textbf{Product Manager}: The Product Manager is responsible for understanding the needs of the users and defining the product's features accordingly. It can use a plugin that collects and analyzes user feedback to understand what features are most important to the users. This agent can also interact with the User Experience Designer Agent to ensure that the product's design aligns with the users' needs.
    \item \textbf{Software Architect}: The Software Architect Agent is responsible for designing the software's architecture. It can use a plugin that simulates different architectural designs to test their performance and scalability. This agent can also interact with the Software Developer Agent to ensure that the architecture is implemented correctly.
    \item \textbf{Software Developer}: The Software Developer is responsible for writing the code that implements the software's features. It can use a plugin that provides access to a code repository to store and manage the code. This agent can also interact with the Software Architect Agent to ensure that the code aligns with the architecture.
    \item \textbf{Software Tester}: The Software Tester is responsible for testing the software to ensure that it works correctly. It can use a plugin that automates the testing process, running a suite of tests on the code and reporting any failures. This agent can also interact with the Software Developer Agent to help identify and fix any bugs in the code.
    \item \textbf{User Interface Designer}: The User Interface Designer is responsible for designing the software's user interface. It can use a plugin that simulates user interactions to test different designs and gather data on user preferences. This agent can then use this data to refine the design.
    \item \textbf{Debugger}: The Debugger is responsible for identifying and fixing bugs in the code. It can use a plugin that provides debugging tools to help identify the cause of any bugs. This agent can also interact with the Software Developer Agent to help fix the bugs.
    \item \textbf{Oracle Agent}: The oracle agent in this context can be used to provide feedback on the overall software development process. It can assess the performance of the other agents and provide feedback to help them improve. For example, it might suggest that the Software Developer Agent needs to write more efficient code, or that the User Experience Designer Agent needs to consider a different design approach.
\end{itemize}

In this way, our model can be used to utilize a more efficient and effective software development process. By assigning specific roles to each agent and using plugins to enhance their capabilities, we can create a system that is capable of automating the development process of a high-quality software, based on the user's needs.

\section{Challenges and Limitations}
\label{challenges}

Multi-agent systems, by their very nature, are complex entities. They involve the interaction of multiple autonomous agents, each with its own capabilities and objectives. This complexity, while being a source of the system's strength, also gives rise to a host of challenges and limitations. In the following subsections, we will explore some of these challenges, shedding light on the potential hurdles that need to be overcome in the context of multi-agent systems.

\subsection{Challenges of a Dynamic System}

The dynamic addition of agents, while offering the potential for enhanced flexibility and adaptability, also presents several challenges. One of the primary concerns is the risk of over-proliferation of agents, which could lead to resource exhaustion or inefficiencies in the system. To mitigate this risk, the system needs to incorporate mechanisms to monitor and control the creation of new agents.

Specifically, the system needs to employ a resource management module that tracks the computational resources consumed by each agent and the system as a whole. This module can alert the system when resource usage approaches a predefined threshold, triggering measures to prevent resource exhaustion. These measures could include halting the creation of new agents.

In addition to resource management, the system also needs to ensure that the dynamic addition of agents does not lead to inefficiencies or conflicts. This is achieved through a coordination mechanism that oversees the assignment of roles and tasks to the agents. When a new agent is created, this mechanism ensures that its role and tasks do not overlap significantly with those of existing agents, thereby preventing redundancies and potential conflicts.

\subsection{Scalability}

Scalability is another significant challenge in multi-agent systems. As the system grows in size and complexity, maintaining the performance and efficiency of the system can become increasingly difficult. The computational resources required for managing the interactions and operations of a large number of agents can be substantial. Additionally, as the number of agents increases, the potential for conflicts and inconsistencies also increases, which can further impact the performance of the system.

\subsection{System Evaluation}

Evaluating the performance of a multi-agent system can be challenging due to the complexity and diversity of the tasks that the system can handle. Traditional evaluation metrics might not be sufficient or appropriate for assessing the performance of the system. Therefore, new evaluation metrics and methodologies might need to be developed to accurately measure the performance of the system and its individual agents.

\subsection{Ethical Considerations}

The use of multi-agent systems also raises several ethical considerations. For instance, the system might make decisions or take actions that have significant impacts on individuals or society. Therefore, it is crucial to ensure that the system operates in an ethical manner and that it respects the rights and interests of all users. This requires careful design and oversight of the system, as well as the implementation of appropriate ethical guidelines and safeguards.

\section{Conclusion}
\label{conclusion}
In this paper, we proposed a novel framework for enhancing the performance and capabilities of LLMs by leveraging the power of multi-agent systems. Our framework introduces a black box environment where multiple IGAs, each with unique attributes and roles, collaborate to handle complex tasks more efficiently and effectively. By introducing collaboration and knowledge exchange among these agents, our system seeks to push the boundaries of what AI can achieve, potentially paving the way towards achieving a higher level of AGI.

Despite the potential benefits, the proposed framework also presents several challenges and limitations, including issues related to security and privacy, agent design and training, system evaluation, and ethical considerations. Addressing these challenges will require further research and development, as well as careful consideration of the ethical implications of deploying such systems. Another promising direction for future work could involve the use of the proposed framework to specific use cases or domains. For instance, the framework could be adapted to handle complex tasks in areas such as healthcare, finance, education, or transportation. This could provide valuable insights into the practical utility and potential impact of the proposed framework.

\bibliographystyle{unsrt}
\bibliography{main}

\end{document}